\documentclass[12pt,onecolumn]{IEEEtran}
\usepackage{setspace}
\usepackage[section]{placeins}
\usepackage{graphicx,subfigure}
\usepackage{amsmath}
\usepackage{amsfonts}
\usepackage{caption2}
\usepackage{amssymb}
\usepackage{varwidth}
\usepackage[rightbars,color]{changebar}
\usepackage{epsfig}
\usepackage{multirow}
\usepackage{booktabs}
\usepackage[linesnumbered,ruled,vlined]{algorithm2e}
\usepackage{array}
\usepackage{rotating}
\usepackage{epsfig,amsfonts,multirow}
\usepackage{changebar}
\newcommand{\punt}[1]{}
\begin{document}
\title{Graph Regularized Low Rank Representation for Aerosol Optical Depth Retrieval}
\author{Yubao~Sun, ~Renlong~Hang, ~Qingshan~Liu, ~Fuping~Zhu, ~Hucheng~Pei
\thanks{Y. Sun, R. Hang and Q. Liu are with the Jiangsu Key Laboratory of Big Data Analysis Technology,
the School of Information and Control, Nanjing University of Information Science and Technology,
Nanjing 210044, China (e-mail:  sunyb$@$nuist.edu.cn; renlong\_hang$@$163.com; qsliu$@$nuist.edu.cn).

F. Zhu and H. Pei are with Beijing Electro-Mechanical Engineering Institute, Beijing 100083, China.

Corresponding author: Qingshan Liu}}

\maketitle

\begin{spacing}{2}
\begin{abstract}
In this paper, we propose a novel data-driven regression model for aerosol optical depth (AOD) retrieval. First, we adopt a low rank representation (LRR) model to learn a powerful representation of the spectral response. Then, graph regularization is incorporated into the LRR model to capture the local structure information and the nonlinear property of the remote-sensing data. Since it is easy to acquire the rich satellite-retrieval results, we use them as a baseline to construct the graph. Finally, the learned feature representation is feeded into support vector machine (SVM) to retrieve AOD. Experiments are conducted on two widely used data sets acquired by different sensors, and the experimental results show that the proposed method can achieve superior performance compared to the physical models and other state-of-the-art empirical models.

\end{abstract}
\IEEEpeerreviewmaketitle

\section{Introduction}\label{Introduction}
Aerosols are small solid and liquid particles suspended in the atmosphere. They can scatter and absorb solar radiance, and modify microphysical and radiative properties of cloud \cite{kaufman2002}. An important metric of aerosols' concentration in the atmosphere is aerosol optical depth (AOD), which measures the amount of depletion that a beam of solar radiation undergoes as it passes through the atmosphere. AOD has become a major atmospheric data product derived from various earth observation satellites, such as the Moderate Resolution Imaging Spectroradiometer (MODIS) \cite{remer2005}, the Multiangle Imaging Spectro-Radiometer (MISR) \cite{kahn2005}, etc. These remote sensing satellites obtain geolocated and calibrated radiances, and then retrieval algorithms are used to derive the corresponding AOD values.

Most of the current operating satellite-retrieval algorithms are based on physical models \cite{kaufman1997}\cite{martonchik1998}\cite{levy2007}. These models need to take into account numerous physical variables affecting the radiometric characteristics of remote sensing data, such as atmospheric conditions, solar azimuth and zenith angles, sensor azimuth and zenith angles, etc. Complex mathematical formulations are set up to represent the relationships between these variables according to radiation transfer equation. To simplify the radiative transfer calculations, a lookup table (LUT) is used to  simulate the radiative properties of the atmosphere calculated for expected aerosol types at particular wavelengths, angles and aerosol loading. Spectral reflectance from the LUT is then compared with the satellite-observation value to find the best match, and the corresponding AOT is the final retrieval result. However, due to the complex Earth-atmosphere interaction, it is difficult to consider all related physical variables and to accurately formulate their relationships. Besides, searching LUT is very time consuming.

An efficient alternative is empirical models \cite{nguyen2011}\cite{ristovski2012}. These models can be considered as data-driven regression approaches. First, using the collocated satellite and ground-based observations to train a regression model. Then, the trained model is used to predict AOD for satellite observations without ground-truth. Among these models, neural networks (NN) \cite{ristovski2012}\cite{vucetic2008} and support vector machines (SVMs) \cite{nguyen2011}\cite{lary2009} are two popular methods, because they can approximate the complex non-linear relationships between satellite-based observations and ground-based observations. In contrast to physical models, empirical models don't make \textit{a prior} assumptions on variable relations or rigid functional forms, they directly rely on the available data. In addition, they are computationally less expensive, flexible to different retrieval scenarios and more accurate than the physical models, if sufficient amounts of training data are available. However, in previous works, most empirical models directly exploit a part of or the whole spectral values as features. Since remote-sensing data often suffer from various annoying degradations, e.g., noise contamination, stripe corruption, and missing data, due to the sensor, photon effects, and calibration error \cite{liu2012}\cite{zhang2012}, directly using such corrupted data without any preprocessing may degrade the performance of empirical models.

In this paper, we propose a graph regularized low rank representation model to address the above issues. Low rank representation (LRR), which was successfully employed in natural image denoising \cite{liu2010}\cite{liu2013}, has shown its potential in remote-sensing data analysis \cite{zhang2014}. By stacking the remote-sensing data into a 2-D matrix, it should be low rank due to the high correlations of the spectral information. Moreover, as discussed in \cite{waters2011}, the rank of the matrix constructed by remote-sensing data is bounded by the small number of pure spectral endmembers. We therefore propose to employ LRR to learn a new feature representation for the remote-sensing data.
Besides, due to the effects of multipath scattering, variations in sun-canopy-sensor geometry, nonhomogeneous composition of pixels, and attenuating properties of media, remote-sensing data are often nonlinearly distributed \cite{bachmann2005}. To preserve such nonlinear structure, motivated by \cite{zheng2011}\cite{gao2013}\cite{cai2011}, graph regularization is incorporated into the LRR model.
Nowadays, it is easy to acquire the rich satellite-retrieval results based on physical models. Many researchers attempted to employ such information as features for regression models \cite{nguyen2011}\cite{vucetic2008}\cite{albayrak2013}. Different from them, we use the satellite-retrieval results as a baseline to construct the graph.
This prior information can make the proposed model combine the merits of physical models and regression models to some extent.

The rest of this paper is outlined as follows. Section$~$\ref{method} introduces the proposed method in detail. Section III presents the data sets used and the experimental results, followed by the conclusion in Section IV.

\section{Methodology}\label{method}
\subsection{Representation model}
Assume the remote-sensing data can be represented as a two-dimensional matrix
$\mathbf{X} = [\mathbf{x}_{1},\cdots,\mathbf{x}_{i},\cdots,\mathbf{x}_{n}]\in \mathbf{R}^{\textit{d}\times \textit{n}}$ by stacking the pixels in the original $d-$dimensional spectral feature space as the columns. Since the remote-sensing data are often corrupted by noise in the acquisition process, $\mathbf{X}$ can be written as $\mathbf{X} = \mathbf{X}_{0} + \mathbf{E}$, where $\mathbf{X}_{0}\in \mathbf{R}^{d\times n}$ is the ideal clean data, $\mathbf{E}\in \mathbf{R}^{d\times n}$ is the noise or outlier. Each pixel of $\mathbf{X}_0$ can be represented by a linear combination of the bases in a `dictionary' $\mathbf{A} = [\mathbf{a}_{1},\mathbf{a}_{2},\cdots,\mathbf{a}_{m}]\in \mathbf{R}^{d\times m}$. So we have $\mathbf{X} = \mathbf{AZ} + \mathbf{E}$, where $\mathbf{Z} = [\mathbf{z}_{1},\cdots,\mathbf{z}_{i},\cdots,\mathbf{z}_{n}]\in \mathbf{R}^{m\times n}$ is the coefficient matrix and each $\mathbf{z}_{i}$ corresponds to the new characterization of $\mathbf{x}_{i}$. The dictionary is often overcomplete, and the clean remote-sensing data often lie in a low rank feature space \cite{waters2011}. Thus, we can use an alternative scheme of $\mathbf{A} = \mathbf{X}$ as in \cite{liu2010}\cite{liu2011}, and the purpose becomes to search for the lowest rank solution of $\mathbf{Z}$ by
\begin{equation}\label{LRR_2}
\begin{aligned}
  \min_{\mathbf{Z,E}}&\quad\|\mathbf{Z}\|_{*} + \lambda\|\mathbf{E}\|_{2,1}, \\
  s.t. &\quad \mathbf{X = XZ + E},
\end{aligned}
\end{equation}
where $\|\cdot\|_{*}$ denotes the \textit{nuclear norm} of a matrix, and $\mathbf{E}$ represents the sparse noise,  which is measured by $l_{2,1}$ norm, i.e., $\|\mathbf{E}\|_{2,1} = \sum_{j=1}^{n}\sqrt{\sum_{i=1}^{d}(\mathbf{E}_{ij})^{2}}$ as in \cite{liu2013}. $\lambda >0$ is a balance parameter.  This is a popular LRR model. It was demonstrated that LRR is capable of capturing the global structure of the data as well as robust to noise \cite{liu2011}.

\begin{figure}
  \centering
  \includegraphics[scale = 0.35]{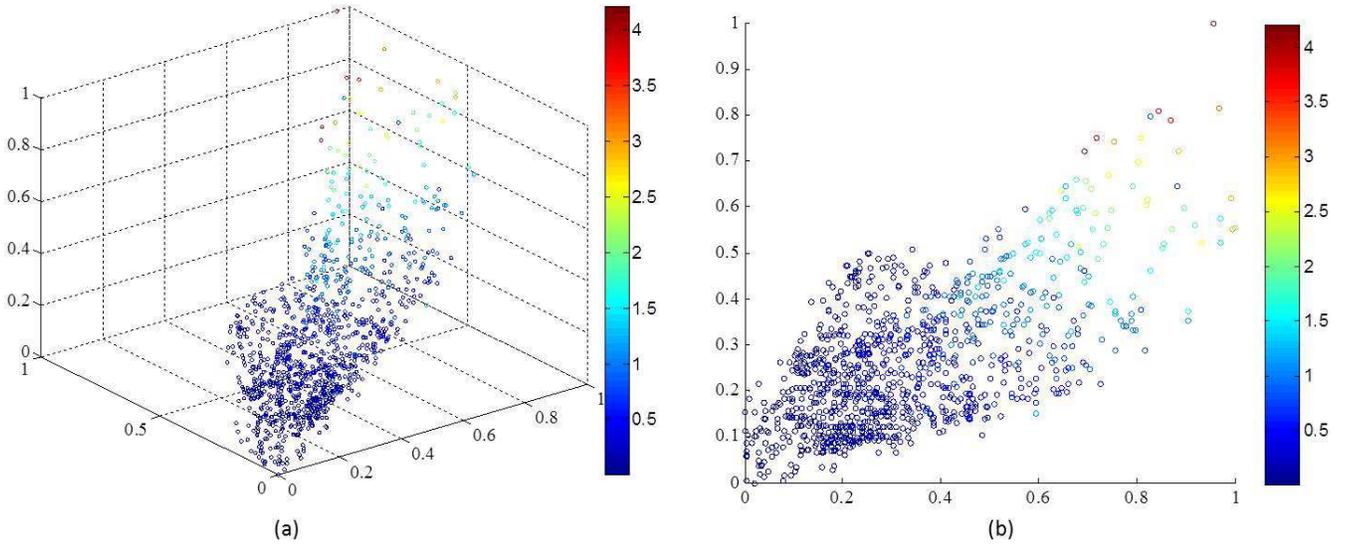}\\
  \caption{An example of nonlinear distribution of MISR remote-sensing data. (a) A three-dimensional map by choosing three spectral bands in MISR. (b) A two-dimensional map of (a). }\label{nonlinear}
\end{figure}

However, as discussed in \cite{lunga2014}, nonlinearities often exist in remote-sensing data due to the effects of many factors. Taking MISR remote-sensing data as an example in Fig.$~$\ref{nonlinear}. Fig.$~$\ref{nonlinear}(a) shows a three-dimensional map by choosing three spectral bands in MISR.
Obviously, this data is nonlinear distribution, which can also be demonstrated from its two-dimensional map. Recently, nonlinear manifold learning has been proved to be a successful method to capture such nonlinearity \cite{ma2010}\cite{li2012}. Inspired by the idea, we propose a graph regularization based LRR model to preserve the local neighboring relations among the data. Thus, the objective function in Eq.$~$(\ref{LRR_2}) can be reformulated as
\begin{equation}\label{LRR_3}
\begin{aligned}
  \min_{\mathbf{Z,E}}\quad\|\mathbf{Z}\|_{*} & + \lambda\|\mathbf{E}\|_{2,1} + \frac{\beta}{2}tr(\mathbf{ZLZ}^{\top}),\\
  s.t. \quad & \mathbf{X  = XZ + E}.
\end{aligned}
\end{equation}
 The term of $Tr(\mathbf{ZLZ}^{\top})$ is the graph regularization term, which is derived as follows:
\begin{equation}\label{graph_regularization}
  \begin{aligned}
    \min_{\mathbf{Z}}& \quad\sum_{i=1}^{n}\sum_{j=1}^{n}\|\mathbf{z}_{i}-\mathbf{z}_{j}\|^{2}w_{ij} \\
    = &\,\min_{\mathbf{Z}} Tr(\mathbf{Z(D-W)Z}^{\top}),\\
    = &\min_{\mathbf{Z}} Tr(\mathbf{ZLZ}^{\top}),
  \end{aligned}
\end{equation}
where $\mathbf{W}$ is an affinity matrix, $\mathbf{D}$ is a diagonal matrix whose elements equal to the sum of rows or columns of $\mathbf{W}$, i.e., $\mathbf{D}_{ii} = \sum_{j=1}^{n}w_{ij}$, and $\mathbf{L = D - W}$ is the graph Laplacian matrix.

An intuitive motivation behind Eq.$~$(\ref{graph_regularization}) is that if two pixels are close, their new representations are also close to each other \cite{belkin2001}. As we all know, it is easy
to acquire the rich satellite-retrieval results, we therefore use them as a baseline to construct the graph. Assume that the satellite-retrieval results of two pixels $\mathbf{x}_{i}$ and $\mathbf{x}_{j}$ are $y_{i}$ and $y_{j}$, respectively, each element $w_{ij}$ of $\mathbf{W}$ can be calculated as
\begin{align}\label{similarity}
    w_{ij} = \left\{
   \begin{array}{ll}
    e^{-\frac{(y_{i}-y_{j})^{2}}{2\sigma^{2}}}, & \hbox{if}\: \;y_{i}\in N_{k}(y_{j}) \;or \;y_{j}\in N_{k}(y_{i}),\\
    \quad0, & \hbox{otherwise},
   \end{array}
  \right.
\end{align}
where $N_{k}(y_{j})$ denotes the set of $k$-nearest neighbors of $y_{j}$, and $\sigma$ refers to the parameter of the Heat kernel. This weight setting promotes our model to be consistent with the satellite-retrieval results, which can inherit the merits of both physical model and regression model.

\subsection{Optimization algorithm}
The objective function Eq.$~$(\ref{LRR_3}) of the graph regularized LRR is non-convex, thus jointly optimizing $\mathbf{Z}$ and $\mathbf{E}$ is extremely difficult. As in \cite{lin2011}\cite{yang2013}, we adopt the linearized Alternating Direction Method of Multipliers (ADMM) algorithm to optimize it. We first convert $~$(\ref{LRR_3}) to the following equivalent problem:
\begin{equation}
\begin{aligned}
  \min_{\mathbf{Z,E,J}}\quad\|\mathbf{J}\|_{*} & + \lambda\|\mathbf{E}\|_{2,1} + \frac{\beta}{2}tr(\mathbf{ZLZ}^{\top}),\\
  s.t. \quad & \mathbf{X  = XZ + E} , \quad \mathbf{Z = J},
\end{aligned}
\end{equation}
which can be changed to the Augmented Lagrange Multiplier (ALM) problem:
\begin{equation}\label{ALM}
\begin{aligned}
  \min_{\mathbf{Z,E,J}}\quad\|\mathbf{J}\|_{*} + \lambda\|\mathbf{E}\|_{2,1} + \frac{\beta}{2}tr(\mathbf{ZLZ}^{\top}) + \langle \mathbf{Y}_{1}, \mathbf{Z-J}\rangle + \\ \langle \mathbf{Y}_{2}, \mathbf{XZ + E - X}\rangle
  + \frac{\mu}{2}\|\mathbf{XZ + E - X}\|_{F}^{2} +\frac{\mu}{2}\|\mathbf{Z - J}\|_{F}^{2},
\end{aligned}
\end{equation}
where $\mathbf{Y}_{1}$, $\mathbf{Y}_2$ are Lagrange multipliers, $\mu > 0$ is a penalty parameter, and $\|\cdot\|_{F}$ denotes the Frobenius norm. Eq.$~$(\ref{ALM}) can be rewritten as
\begin{equation}\label{ALM_2}
\begin{aligned}
  \min_{\mathbf{Z,E,J}}\quad\|\mathbf{J}\|_{*} + \lambda\|\mathbf{E}\|_{2,1} + \frac{\beta}{2}tr(\mathbf{ZLZ}^{\top}) +
  \frac{\mu}{2}\|\mathbf{XZ + E - X} + \frac{\mathbf{Y}_{2}}{\mu}\|_{F}^{2} + \frac{\mu}{2}\|\mathbf{Z - J +} \frac{\mathbf{Y}_{1}}{\mu}\|_{F}^{2}.
\end{aligned}
\end{equation}
Following the iterative optimization method in \cite{liu2010}, we divide Eq.$~$(\ref{ALM_2}) into three sub-problems: optimizing $\mathbf{J}$ while fixing $\mathbf{Z}$ and $\mathbf{E}$, optimizing $\mathbf{E}$ while fixing $\mathbf{J}$ and $\mathbf{Z}$, and optimizing $\mathbf{Z}$ while fixing $\mathbf{J}$ and $\mathbf{E}$.

\textit{Fixing $\mathbf{Z}$ and $\mathbf{E}$ to optimize $\mathbf{J}$,} Eq.$~$(\ref{ALM_2}) is simplified to
\begin{equation}\label{sub-problem1}
 \min_{\mathbf{J}}\|\mathbf{J}\|_{*} + \frac{\mu}{2}\|\mathbf{J}-(\mathbf{Z} + \frac{\mathbf{Y}_{1}}{\mu})\|_{F}^{2}.
\end{equation}
According to \cite{lin2010} \cite{cai2010}, singular value thresholding operator is used to solve
Eq.$~$(\ref{sub-problem1}).

\textit{Fixing $\mathbf{J}$ and $\mathbf{Z}$ to optimize $\mathbf{E}$,} Eq.$~$(\ref{ALM_2}) is reduced to
\begin{equation}\label{sub-problem2}
  \min_{\mathbf{E}}\lambda\|\mathbf{E}\|_{2,1} + \frac{\mu}{2}\|\mathbf{E} - (\mathbf{X - XZ - } \frac{\mathbf{Y}_{2}}{\mu})\|_{F}^{2}.
\end{equation}
According to the \textbf{Lemma 3.2} in \cite{liu2010}, if the optimal solution
is $\mathbf{E}^{*}$, the $i-th$ column of $\mathbf{E}^{*}$ is:
\begin{align}
    \mathbf{E}^{*}(:,i) = \left\{
   \begin{array}{ll}
    \frac{\|\mathbf{q}_{i}\| - \frac{\lambda}{\mu}}{\|\mathbf{q}_{i}\|}\mathbf{q}_{i}, & \hbox{\textit{if}}\: \frac{\lambda}{\mu}<\|\mathbf{q}_{i}\|, \\
    \quad0, & \hbox{otherwise}.
   \end{array}
  \right.
  \end{align}
where $\mathbf{q}_{i}$ is the $i-th$ column vector of matrix $\mathbf{X - XZ }- \frac{\mathbf{Y}_{2}}{\mu}$.

\textit{Fixing $\mathbf{J}$ and $\mathbf{E}$ to optimize $\mathbf{Z}$,} Eq.$~$(\ref{ALM_2}) is simplified to
\begin{equation}\label{sub_problem3}
\min_{\mathbf{Z}}\frac{\beta}{2}tr(\mathbf{ZLZ}^{\top})+\frac{\mu}{2}\|\mathbf{XZ+E-X}
+\frac{\mathbf{Y}_{2}}{\mu}\|_{F}^{2}+\frac{\mu}{2}\|\mathbf{Z-J}+\frac{\mathbf{Y}_{1}}{\mu}\|_{F}^{2}.
\end{equation}
We adopt a linearization strategy to optimize Eq.$~$(\ref{sub_problem3}). In specific,  $\frac{\mu}{2}\|\mathbf{XZ + E - X} + \frac{\mathbf{Y}_{2}}{\mu}\|_{F}^{2}$ is linearly approximated into the following formula by using second-order Taylor expansion around point $\mathbf{Z}_{k}$:
\begin{equation}\label{Taylor}
\frac{\mu}{2}\|\mathbf{XZ + E - X} + \frac{\mathbf{Y}_{2}}{\mu}\|_{F}^{2}\approx
\langle\mu \mathbf{A}^{\top}(\mathbf{XZ}_{k}+\mathbf{E-X}+\frac{\mathbf{Y}_{2}}{\mu}),\mathbf{Z-Z}_{k}\rangle+\frac{\mu\|\mathbf{A}\|_{F}^{2}}{2}\|\mathbf{Z-Z}_{k}\|_{F}^{2}.
\end{equation}
Then, substitute (\ref{Taylor}) for $\frac{\mu}{2}\|\mathbf{XZ + E - X} + \frac{\mathbf{Y}_{2}}{\mu}\|_{F}^{2}$ and Eq.$~$(\ref{sub_problem3}) is written as follows:
\begin{equation}\label{Final}
\min_{\mathbf{Z}}\frac{\beta}{2}tr(\mathbf{ZLZ}^{\top})+\frac{\mu\|\mathbf{X}\|_{F}^{2}}{2}\|\mathbf{Z}-(\mathbf{Z}_{k}-\frac{\mathbf{X}^{\top}(\mathbf{XZ}_{k}
+\mathbf{E-X}+\frac{\mathbf{Y}_{2}}{\mu})}{\|\mathbf{X}\|_{F}^{2}})\|_{F}^{2}+\frac{\mu}{2}\|\mathbf{Z-J}+\frac{\mathbf{Y}_{1}}{\mu}\|_{F}^{2}.
\end{equation}
Finally, we can achieve the optimal $\mathbf{Z}$ by setting the derivative of Eq.$~$(\ref{Final}) with respect to $\mathbf{Z}$ to zero:
\begin{equation}
\mathbf{Z}=[\mu\|\mathbf{X}\|_{F}^{2}\mathbf{Z}_{k}-\mu \mathbf{X}^{\top}\mathbf{XZ}_{k}-\mu \mathbf{X}^{\top}\mathbf{E}+\mu \mathbf{X}^{\top}\mathbf{X-X}^{\top}\mathbf{Y}_{2}+\mu \mathbf{J-Y}_{1}](\beta \mathbf{L}+\mu(\|\mathbf{X}\|_{F}^{2}+1)\mathbf{I})^{-1}.
\end{equation}

The detailed optimization algorithm is summarized in \textbf{Algorithm \ref{algorithm}}.
\begin{algorithm}\label{algorithm}
\DontPrintSemicolon
\KwIn{Data matrix $\mathbf{X}$, parameter $\lambda$ and $\beta$. \\
     \textbf{Initialize:} $\mathbf{Z=J=0}$, $\mathbf{E=0}$, $\mathbf{Y}_{1}=\mathbf{0}$, $\mathbf{Y}_{2}=\mathbf{0}$, $\mu=10^{-6}$,
     $max_{\mu}=10^{11}$, $\rho=1.1$, $\varepsilon=10^{-11}$, $\mathbf{Z}_{k}=0$.}
\KwOut{Lowest rank representation $\mathbf{Z}$.}
\For{each iteration}
{Fix $\mathbf{Z}$, $\mathbf{E}$ and update $\mathbf{J}$:
$\mathbf{J}$ = $argmin\;\|\mathbf{J}\|_{*}+\frac{\mu}{2}\|\mathbf{J}-(\mathbf{Z}+\frac{\mathbf{Y}_{1}}{\mu})\|_{F}^{2}$.\\
Fix $\mathbf{J}$, $\mathbf{Z}$ and update $\mathbf{E}$:
$\mathbf{E}$ = $argmin\;\lambda\|\mathbf{E}\|_{2,1}+\frac{\mu}{2}\|\mathbf{E}-(\mathbf{X-XZ}-\frac{\mathbf{Y}_{2}}{\mu})\|_{F}^{2}$.\\
Fix $\mathbf{J}$, $\mathbf{E}$ and update $\mathbf{Z}$:
$\mathbf{Z}$ = $[\mu\|\mathbf{X}\|_{F}^{2}\mathbf{Z}_{k}-\mu \mathbf{X}^{\top}\mathbf{XZ}_{k}-\mu \mathbf{X}^{\top}\mathbf{E}+\mu \mathbf{X}^{\top}\mathbf{X}-\mathbf{X}^{\top}\mathbf{Y}_{2}+\mu \mathbf{J-Y}_{1}](\beta \mathbf{L}+\mu(\|\mathbf{X}\|_{F}^{2}+1)\mathbf{I})^{-1}$.\\
Update the multipliers:
$\mathbf{Y}_{1} = \mathbf{Y}_{1}+\mu(\mathbf{Z-J})$; $\mathbf{Y}_{2} = \mathbf{Y}_{2}+\mu(\mathbf{XZ+E-X})$.\\
Update the parameters:
$\mu = \min(\rho\mu, \max_{\mu})$; $\mathbf{Z}_{k}=\mathbf{Z}$.\\
Check the convergence conditions:
$\|\mathbf{X-XZ-E}\|_{\infty}<\varepsilon$ and $\|\mathbf{Z-J}\|_{\infty}<\varepsilon$.
}
\caption{The proposed optimization algorithm for graph regularized LRR by linearized ADMM}
\end{algorithm}

\section{Experiments}
\subsection{Data Sets}
As in \cite{ristovski2012}\cite{vucetic2008}\cite{lary2009}, we use Level 2.0 AERONET retrievals as the target values for regression models. AERONET is a global network of about 250 ground-based instruments that observe aerosols \cite{holben1998}. Most of these stations measure AOD in different spectral bands centered around the nominal wavelengths of 340, 380, 440, 670 $nm$, and others \cite{petrenko2012}. To facilitate inter-comparisons with other instruments, these data are interpolated to 550 $nm$ using the quadratic fit on log-log scale from all wavelengths, at a particular location and time \cite{remer2005}. Besides using the AOT values at 550 $nm$ as ground-truth of regression problems, the following two sensors' spectral values along all the bands are used as features (inputs).

The first is the Moderate Resolution Imaging Spectroradiometer (MODIS) data. MODIS is a key instrument aboard the TERRA satellite for the collection of aerosol and cloud information. It has a swath width of 2330 $km$,  and achieves a global coverage in about two days. The MODIS instrument has a single camera observing the top-of-the-atmosphere reflectance over 36 spectral bands between 410 $nm$ and 14 $\mu m$ at three different spatial resolutions (250 $m$, 500 $m$, 1 $km$) \cite{salomonson1989}. We obtain the MODIS Level-1B calibrated radiance product MOD021KM with spatial resolution of 1 $km$, covering the Beijing AERONET location between January 2002 and December 2014. Over the same spatial and temporal range, we obtain the Level-2 aerosol-retrieval product MOD04 (Collection 6, QA$>$1) with a spatial resolution of 10 $km$, and geolocation product MOD03 with 1 $km$ resolution. MOD04 product is used as a baseline to verify the effectiveness of regression models. Thereafter, Level 2.0 AERONET data are collocated in space and time with the MODIS data. The detailed process can be found in \cite{vucetic2008}. We obtain a total of 843 spatially and temporally collocated observations from MODIS and AERONET.

The second is Multi-angle Imaging SpectroRadiometer (MISR) data. MISR is one of the five instruments mounted on Terra spacecraft. The spacecraft flies in a sun-synchronous 705 $km$ descending polar orbit, so that it crosses the equator always at 10:30 $am$ local time. The MISR instrument consists of nine pushbroom cameras arranged in different view angles relative to the earth's surface. Each camera uses four Charge-Coupled Device (CCD) line arrays in a single focal plane. The line arrays cover 360 $km$ wide swath and provide four spectral bands in Blue, Green, Red and Near Infrared (NIR) that are centered at 443, 555, 670 and 865 $nm$, respectively. The resolution of all the four bands in nadir view and the red band at all the nine angles is 275 $m$ and the resolution of the other bands is 1.1 $km$. We download 1045 collocated MISR and AERONET data from MAPSS \cite{petrenko2012}, covering the whole 23 stations at all available time in China.

\subsection{Experimental Setup}
To demonstrate the superiority of the proposed graph regularized LRR model, the new representation is feeded into the subsequent regression model SVM. For simplicity, we name it GLRR+SVM, which is compared with the following six retrieval models: 1) the operating satellite-retrieval algorithms by physical models; 2) the ordinary least square regression (OLS); 3) ridge regression (RR); 4) NN; 5) SVM; 6) the classical LRR with SVM regressor (LRR+SVM). For NN, the optimal number of hidden nodes is chosen from [2, 50] in steps of 5 via a 5-fold cross validation. For SVM, we adopt the Gaussian kernel since it usually achieves the best results compared to other kernels. The optimal variance parameter $\gamma$ for the Gaussian kernel and the regularization parameter $C$ in SVM are both selected from $\{10^{-3},10^{-2},\cdots,10^{3}\}$. Besides, there are two regularization parameters $\lambda$ and $\beta$ for the graph regularized LRR model, which are also chosen by a 5-fold cross validation from the given set $\{10^{-3},10^{-2},\cdots,10^{3}\}$.
In all the experiments, we randomly divide the whole data into the training set and the testing set according to some percentages. The training set is used to train the regression based retrieval models, while the testing set is used to evaluate their performances. In order to reduce the effects of random selection, all the algorithms are repeated ten times and the average performances are reported. Without loss of generality, we use two mainstream evaluation metrics: the root-mean-square error (RMSE) to evaluate the accuracy of the estimations, and Pearson's correlation coefficient R to evaluate the goodness of fit.
\begin{figure}[htp]
  \centering
  \includegraphics[scale = 0.35]{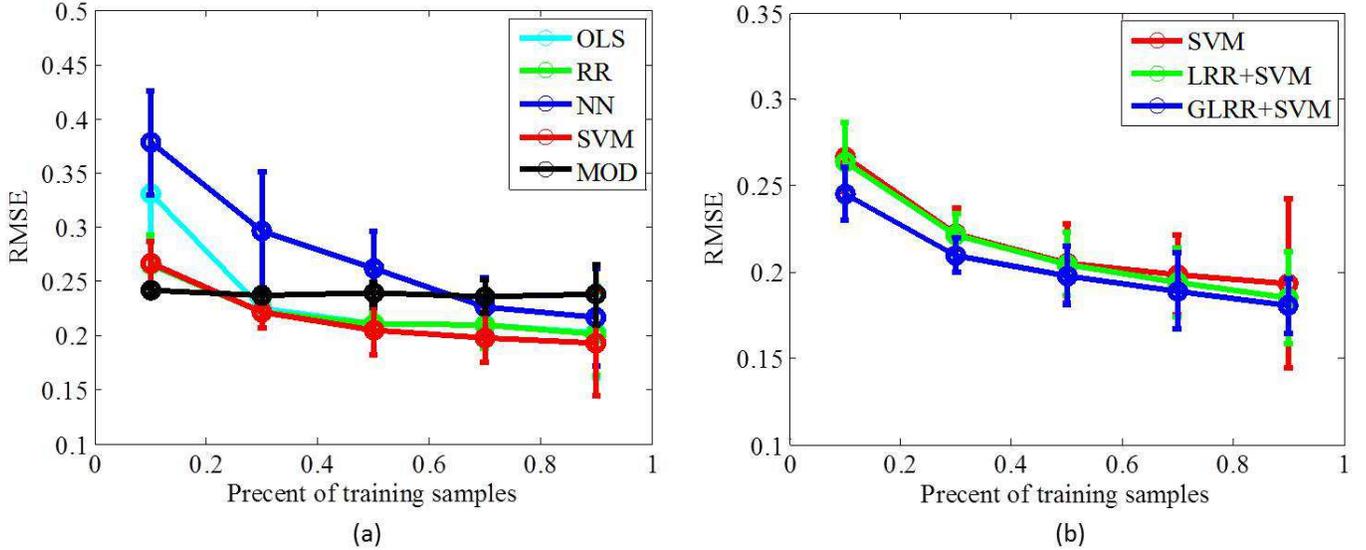}\\

  \caption{RMSE and standard deviations achieved by seven different methods on MODIS data set. (a) Comparisons among SVM and four other methods. (b) Comparisons among SVM, LRR+SVM and GLRR+SVM.}\label{MOD_RMSE}
\end{figure}

\begin{figure}[htp]
  \centering
  \includegraphics[scale = 0.35]{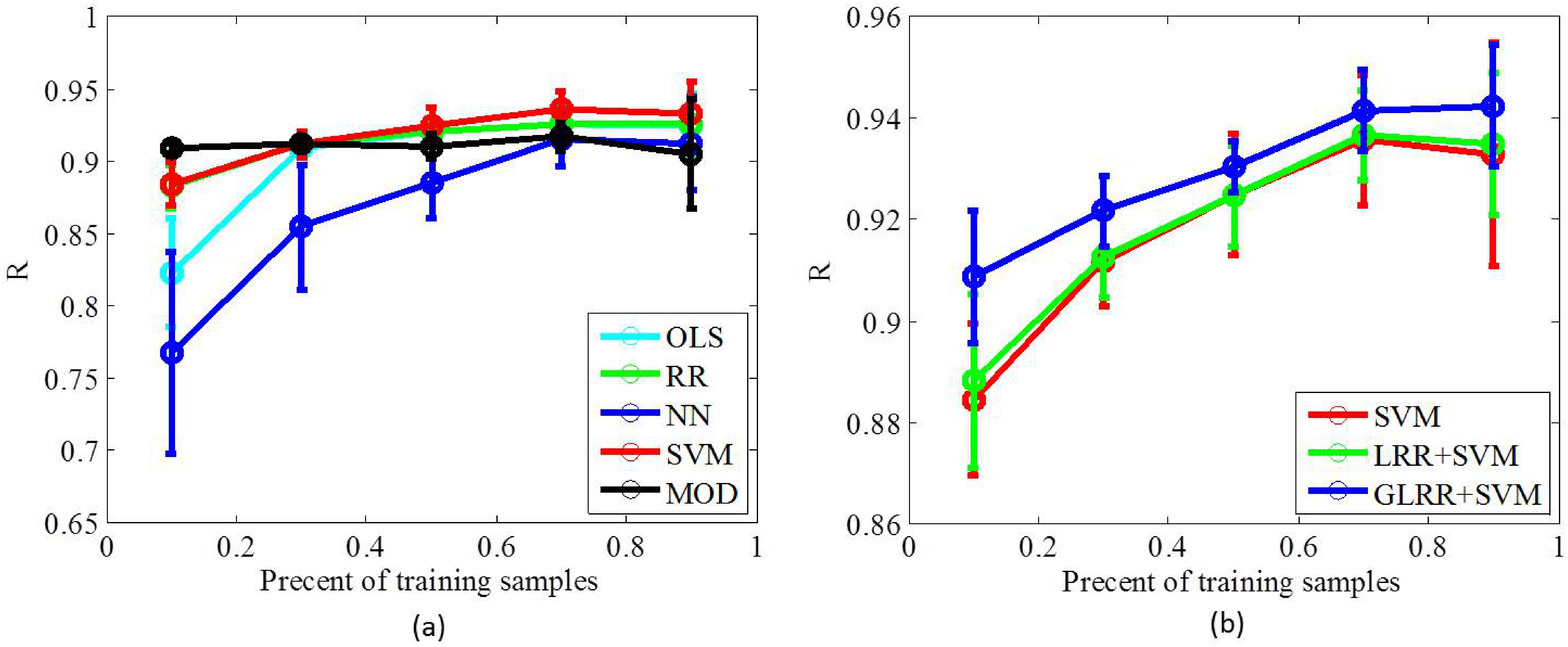}\\
  \caption{R and standard deviations achieved by seven different methods on MODIS data set. (a) Comparisons among SVM and four other methods. (b) Comparisons among SVM, LRR+SVM and GLRR+SVM.}\label{MOD_R}
\end{figure}

\begin{figure}[htp]
  \centering
  \includegraphics[scale = 0.35]{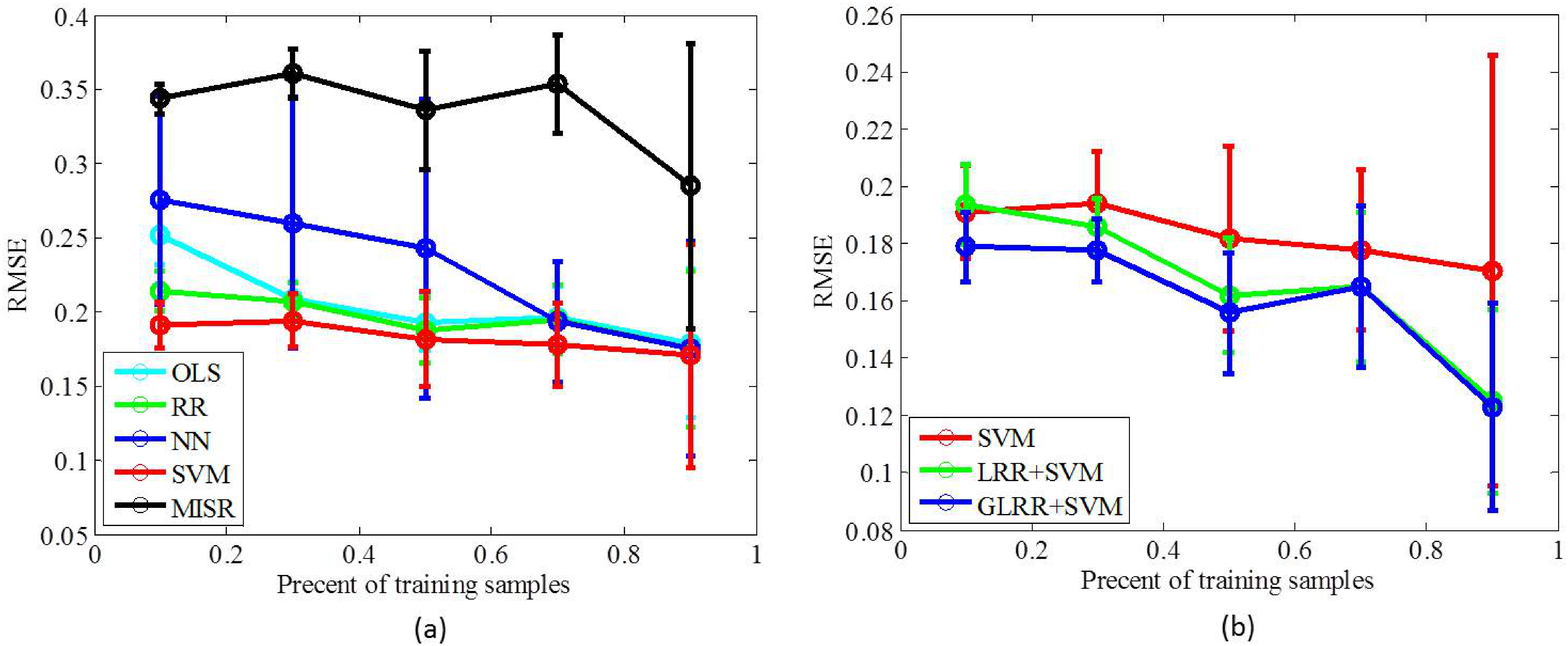}\\
  \caption{RMSE and standard deviations achieved by seven different methods on MISR data set. (a) Comparisons among SVM and four other methods. (b) Comparisons among SVM, LRR+SVM and GLRR+SVM.}\label{MISR_RMSE}
\end{figure}

\begin{figure}[htp]
  \centering
  \includegraphics[scale = 0.35]{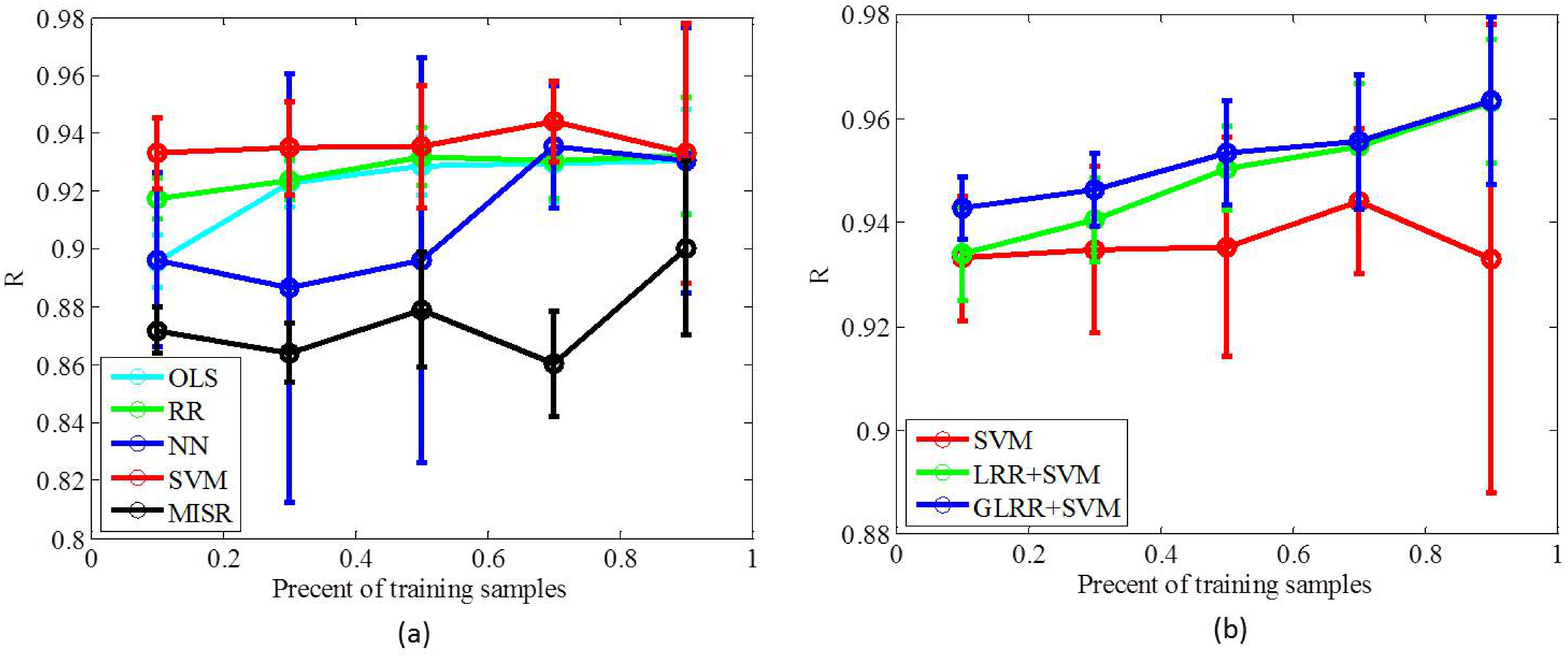}\\
  \caption{R and standard deviations achieved by seven different methods on MISR data set. (a) Comparisons among SVM and four other methods. (b) Comparisons among SVM, LRR+SVM and GLRR+SVM.}\label{MISR_R}
\end{figure}
\subsection{Results and Discussion}

For the MODIS data set, the average RMSE values and the standard deviations achieved by seven different models from ten experiments are demonstrated in Fig.$~$\ref{MOD_RMSE}, where the smaller values correspond to the better performances. Several conclusions can be observed from this figure.
First, in Fig.$~$\ref{MOD_RMSE}(a), it can be seen that given enough training samples, the regression models achieve higher performance than the physical model. Second, SVM yields the best performance in the four regression models, because it can well solve the case with small numbers of training samples.
Third, in Fig.$~$\ref{MOD_RMSE}(b), the performance of LRR+SVM is a little better than SVM especially when the percentage of training samples is more than 70\%. This indicates that LRR is able to learn a better representation from the corrupted observation data compared to the pure spectral response values.
More importantly, GLRR+SVM further improves the performance as compared to LRR+SVM, because it can capture the local structure information and the nonlinear property of MODIS data.
The last but not the least, a deficiency of regression models can be observed in Fig.$~$\ref{MOD_RMSE}(a). Specifically, when the percentage of training samples is 10\%, the physical model is better than all of the regression models. However, when the percentage of training samples is more than 20\%, the best regression model SVM yields higher performance than the physical model. This indicates that the regression models heavily rely on the number of training samples while the physical model is stable. The above conclusions can also be verified from another evaluation indicator R in Fig.$~$\ref{MOD_R}. Different from Fig.$~$\ref{MOD_RMSE}, here, the larger values denote the better performances. In particular, SVM obtains the best results compared to OLS, RR and NN. With a learned representation, LRR+SVM improves the performance of SVM. Besides, GLRR+SVM is capable of boosting the results of LRR+SVM by adding a graph regularization into the original LRR.

For the MISR data set, Fig.$~$\ref{MISR_RMSE} and Fig.$~$\ref{MISR_R} show the average performances and deviations of seven different models from ten experiments. From these figures, we can observe that the performances of the regression models are better than that of the physical model. In particular, SVM is superior to OLS, RR and NN. GLRR+SVM and LRR+SVM are both better than SVM, because they can learn an effective representation rather than directly using the corrupted spectral values. More importantly, when graph regularization based on the physical retrieval results is incorporated into the LRR model, the performance can be further boosted especially when the number of training samples is less than 70\%, which certify the effectiveness of GLRR+SVM.

\begin{figure}[htp]
  \centering
  \includegraphics[scale = 0.35]{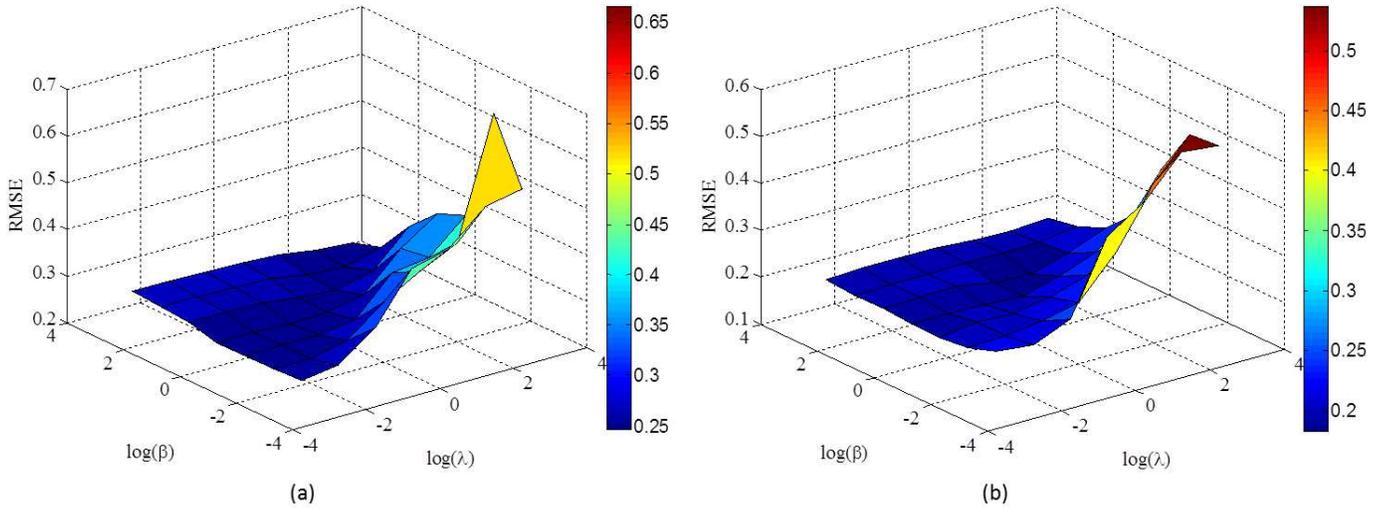}\\
  \caption{RMSE of the proposed method versus two different regularization parameters $\lambda$ and $\beta$ on (a) MODIS data set. (b) MISR data set.}\label{parameter_analysis}
\end{figure}
\subsection{Parameter Analysis}
There are two important regularization parameters $\lambda$ and $\beta$ in the proposed method. The first one is used to balance the effects of low rank property and noise component, and the second one is utilized to balance the information from empirical model and physical model. Fig.$~$\ref{parameter_analysis} demonstrates the 3-D diagram of RMSE against them on MODIS and MISR data sets. From Fig.$~$\ref{parameter_analysis}(a), we can observe that as $\lambda$ and $\beta$ increase, the corresponding RMSE firstly increases and then decreases, and achieves the maximal value at $\lambda = 1$ and $\beta = 10^{-1}$. Similar results can be seen from Fig.$~$\ref{parameter_analysis}(b), and the maximal value appears when $\lambda = \beta = 10$.

\section{Conclusion}
This paper proposed a graph regularized low rank representation (LRR) model to learn an effective feature representation for aerosol optical depth (AOD) retrieval. Based on the observation that remote-sensing data often lie in a low rank subspace, LRR was naturally used to uncover the intrinsic data structure from corrupted or noisy observations. Additionally, to preserve the nonlinear structure in remote-sensing data, graph regularization was added to the LRR model. It is well known that the current operating satellites can generate AOD values by physical models. Thus, the proposed graph model was constructed based on such rich information. By conducting experiments on two data sets collected by different instruments (MODIS and MISR), we compared the proposed method with the physical models and the other state-of-the-art empirical models. The experimental results indicate that the learned representation can improve the retrieval performance.

\bibliography{IEEEfull,LRR}
\bibliographystyle{IEEEbib}
\end{spacing}
\end{document}